\documentclass{article}

\usepackage{microtype}
\usepackage{graphicx}
\usepackage{subfigure}
\usepackage{booktabs} 

\usepackage{hyperref}

\usepackage[accepted]{icml2023}

\usepackage{amsmath}
\usepackage{amssymb}
\usepackage{mathtools}
\usepackage{amsthm}
\usepackage{bbm}
\usepackage{multirow}
\usepackage{makecell}
\usepackage[capitalize,noabbrev]{cleveref}

\theoremstyle{plain}

\theoremstyle{definition}

\theoremstyle{remark}

\usepackage[textsize=tiny]{todonotes}

\begin{document}

\twocolumn[
\icmltitle{Tree-of-Mixed-Thought: Combining Fast and Slow Thinking for Multi-hop Visual Reasoning}



\icmlsetsymbol{equal}{*}
\icmlsetsymbol{corrsponding}{$\dagger$}
\begin{icmlauthorlist}
\icmlauthor{Pengbo Hu$^*$}{ustc-sist}
\icmlauthor{Ji Qi $^*$}{cmss}
\icmlauthor{Xingyu Li}{lg}
\icmlauthor{Hong Li}{shanghaiTech}
\icmlauthor{Xinqi Wang}{ustc-sist}
\icmlauthor{Bing Quan}{cmss}
\icmlauthor{Ruiyu Wang}{cmss}
\icmlauthor{Yi Zhou}{ustc-sist,corrsponding}
\end{icmlauthorlist}

\icmlaffiliation{ustc-sist}{University of Science and Technology of China}
\icmlaffiliation{shanghaiTech}{ShanghaiTech University}
\icmlaffiliation{lg}{Lin Gang Laboratory}
\icmlaffiliation{cmss}{cmss.chinamobile.com}

\icmlcorrespondingauthor{Pengbo Hu}
{pbhu@mail.ustc.edu.cn}

\vskip 0.3in
]


\printAffiliationsAndNotice{\icmlEqualContribution $\dagger$: Corresponding author} 

\begin{abstract}
There emerges a promising trend of using large language models (LLMs) to generate code-like plans for complex inference tasks such as visual reasoning. This paradigm, known as LLM-based planning, provides flexibility in problem solving and endows better interpretability. However, current research is mostly limited to basic scenarios of simple questions that can be straightforward answered in a few inference steps. Planning for the more challenging multi-hop visual reasoning tasks remains under-explored. Specifically, under multi-hop reasoning situations, the trade-off between accuracy and the complexity of plan-searching becomes prominent. The prevailing algorithms either address the efficiency issue by employing the fast one-stop generation or adopt a complex iterative generation method to improve accuracy. Both fail to balance the need for efficiency and performance. Drawing inspiration from the dual system of cognition in the human brain, the fast and the slow think processes, we propose a hierarchical plan-searching algorithm that integrates the one-stop reasoning (fast) and the Tree-of-thought (slow). Our approach succeeds in performance while significantly saving inference steps. Moreover, we repurpose the PTR and the CLEVER datasets, developing a systematic framework for evaluating the performance and efficiency of LLMs-based plan-search algorithms under reasoning tasks at different levels of difficulty. Extensive experiments demonstrate the superiority of our proposed algorithm in terms of performance and efficiency. The dataset and code will be release soon.
\end{abstract}

\begin{figure*}[!t]
    \begin{center}
        \includegraphics[width=0.9\textwidth]{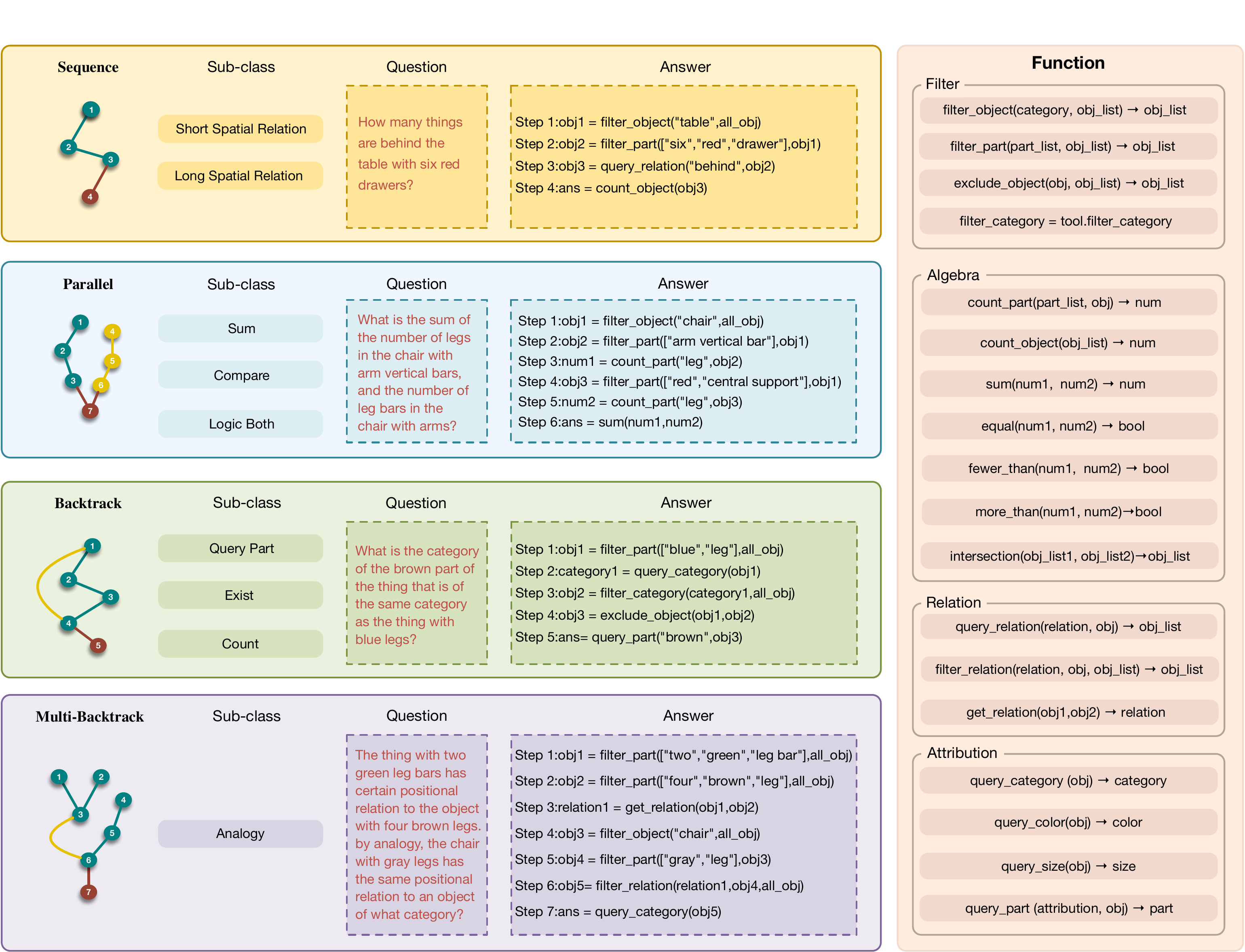}
    \end{center}
    \caption{Dataset Diagram. We repurpose a novel dataset comprising four distinct types of reasoning structures and nine question types. To facilitate seamless interaction with scene graphs, we design dedicated tools. Further details regarding the dataset can be found in the Appendix. }
    \label{fig:dataset}
\end{figure*}

\section{Introduction}

Large Language Models (LLMs) have recently witnessed significant advancements. The most capable and cost-effective model is  ChatGPT~\cite{openai2023chat}, which showcase remarkable capabilities in language comprehension~\cite{bubeck2023sparks,ouyang2022training},reasoning~\cite{wei2022chain,huang2022towards}, plan~\cite{jiang2023self,song2022llm,yao2022react} and tool use~\cite{qin2023toolllm,shen2023hugginggpt,gupta2023visual}. Particularly noteworthy is its capability for generating code-like plans, enabling it to effectively interact external tools through code~\cite{gupta2023visual,shen2023hugginggpt,liang2023taskmatrix}. This capability significantly broadens the scope of applications for language models, thereby enhancing their utility and potential impact~\cite{mialon2023augmented}.

The use of LLMs to generate code-like plain for answering questions has been widely applied in multi-modal tasks~\cite{shen2023hugginggpt,gupta2023visual,wu2023visual}.
These approaches typically involve predefined visual expert models and the use of LLMs to generate plans for scheduling these models, thereby aiding in the resolution of complex visual questions. 
However, empirical investigations have revealed that employing LLMs, such as ChatGPT, for the generation of long-range plans poses significant challenges in achieving stable results, thereby limiting its applicability in complex long-range multi-hop reasoning tasks ~\cite{bubeck2023sparks,bang2023multitask}. To overcome this challenge, the tree-of-thought, which possesses the ability to backtrack and iterate ~\cite{yao2023tree,long2023large}, has been proven to work well on long-range plans generation. Tree-of-thought is typically a 'slow-thinking' or 'system-2' mode of thinking, which can gradually search an optimal planning trajectory through iterative trials and backtracking. 
This enables it to deal with long-range inference problems. 
However, the main limitation of the tree-of-thought is its efficiency, as it requires multiple visits to LLMs to achieve high planning accuracy, resulting in significant time and computational resource consumption.

Given the tree-of-thought's proficient capability in generating long-range plans, it is imperative to explore a methodology that can keep its aptitude while reduce the reasoning step consumption of LLMs.
Recently, some studies have explored the use of LLMs for one-stop generation, often referred to as "fast thinking" or "system-1 thinking"~\cite{bubeck2023sparks}, wherein the LLMs generate the whole plans by only visiting LLMS once.  
However, this approach is often limited by the lack of precision in the generated plans.
Inspired by the 'dual process' of the human brain~\cite{sloman1996empirical}, it is intuitively to combine the ToT (slow thinking) and one-stop (fast thinking) to keep the long-range plans ability while reducing the reasoning steps.
In this paper, we propose two strategies to combine them. Our experiments demonstrate that both strategies significantly save the inference step of LLMs while maintaining accuracy. 

To systematically evaluate our method on long-range multi-hop visual reasoning tasks. There still pose two challenges. First, the existing method ~\cite{shen2023hugginggpt,gupta2023visual,wu2023visual} are constrained by the accuracy of the visual expert models, which limits the ability to effectively evaluate the performance of the language model itself on such tasks. Second, existing methods mainly focus on simple visual questions that require only a few inference steps to answer, thus making it relatively easy for LLMs to generate effective plans. To address these challenges, we identified the use of scene graphs in the PTR~\cite{hong2021ptr} and CLEVR~\cite{johnson2017clevr} datasets. Scene graphs provide a comprehensive representation of image information, which enables us to design tools that can interact with scene graphs. Since these tools can accurately extract visual information from scene graphs, we can effectively evaluate the capabilities of LLMs on their own. In addition, we specifically select different inference structures and varying number of hops in the problem. This allows for an efficient evaluation of the ability of LLMs in long-range planning to solve visual reasoning tasks with different qeustion structures.

Our contributions are three-fold:
\begin{itemize}
\item We propose two strategies to the tree-of-thought(slow thinking) and one-stop geneartion (fast thinking). Our method significantly reduces the re-evaluation step for LLMs while maintaining accuracy.
\item To evaluate the performance of LLMs in multi-hop visual reasoning, we construct a multi-hop reasoning dataset and design a toolkit that enables direct interaction with scene graphs, thereby avoiding errors from the visual expert module. 
\item Through systematic experiments on this dataset, we demonstrate the effectiveness of our approach.    
\end{itemize}
\section{Related work}

\subsection{Plan Generation with LLMs.}
There are three primary paradigms for generating plans using LLMs. First, one-stop generation, wherein LLMs directly generate the entire plan in a single run~\cite{{xie2023translating,vemprala2023chatgpt}}. The second is iterative generation, which generates the plan step by step. Commonly used methods within this paradigm include ReAct~\cite{yao2022react}, tool-former~\cite{schick2023toolformer}, and reflexion~\cite{shinn2023reflexion}. These methods typically allow plan modification through feedback, but cannot explore diverse alternative plan trajectories since these methods cannot backtrack to the pre-state.  Third, tree-of-thought, which incorporates backtracking and iteration features of tree search. This approach leverages the hierarchical structure of the tree to iteratively and recursively search for optimal planning sequences, resulting in planning trajectories that outperform those produced by the previous two approaches. However, this approach normally requires many more visits to the LLMs, resulting in huge time and resource consumption.
In this study, we propose a hybrid approach that combines the strengths of both one-stop and tree-of-thought approaches. Our approach preserves the backtracking and iteration capabilities of tree-of-thought while significantly saving the inference steps of LLMs. 

\subsection{Visual Reasoning with LLMs}
The application of large language models (LLMs) to implement visual reasoning tasks through the generation of code-like plans has been widely explored in various studies, such as VisualChatGPT~\cite{wu2023visual}, HuggingGPT~\cite{shen2023hugginggpt}, and Visual programs~\cite{gupta2023visual}. This approach offers flexibility and interpretability in addressing complex visual inquiries and can be applied to diverse multi-modal tasks. These studies employ an iterative generation process to sequentially generate plans, which restricts their application to relatively straightforward questions that only necessitate a few steps of plan generation. In this study, we place greater emphasis on the challenging long-range multi-hop visual reasoning task. 

\section{Method}

\subsection{Dataset and Evaluation}
To systematically evaluate the capacity of LLMs in addressing multi-hop visual reasoning tasks by generating long-range code-like plans, several factors need to be considered. First, it is crucial to ensure the accuracy of the visual expert model in order to evaluate the capability of LLMs independently. Second, it is necessary to have the distribution of testing questions varying in hop number. Lastly, different reasoning structures for various questions should be incorporated to aid in assessing the planning abilities of LLMs under different inference structures.   

To this end, we repurpose the PTR~\cite{hong2021ptr} and CLEVR~\cite{johnson2017clevr} datasets and develop a new evaluation dataset. Both the PTR and CLEVR datasets are synthetic visual question answering datasets for studying complex multi-hop visual reasoning such as physical reasoning and relational reasoning.  The difference between them is that the question structure of the PTR dataset is more fine-grained than CLEVR, yet CLEVR has questions with long hops. The salient features of both datasets include a diverse range of problem types, encompassing a variety of lengths, hop numbers, reasoning structures, and, most importantly, providing comprehensive scene graphs that completely represent visual information. This enables the design of tools that interact with scene graphs, thus avoiding the problem of inaccurate visual expert models. Consequently, this allows for a more focused evaluation of the performance of LLMs in such tasks, facilitating accurate assessments.

We specifically select four types of inference structures from the PTR and CEVLR datasets that can further categorize the nine types of questions. The details are illustrated in Figure \ref{fig:dataset}. We also equip the dataset with tools that can directly interact with the corresponding scene graphs, most of which are derived from the PTR dataset and Clever dataset.  This ensures that the tools can read the vision information with complete accuracy. 
For each question type, we selected 120 samples, with 20 samples sever as few-shot example library and 100 samples for testing. Detailed descriptions of each question type and its corresponding tools are provided in the Appendix. 

\subsection{Baseline}
\label{sec:baseline}

Given that our approach primarily focuses on integrating one-stop generation and ToT search, we directly utilize these two methods as baselines for comparison and analysis.

\subsubsection{One-Stop} The One-Stop Generation approach stands by a fast-thinking mode. It directly exploits a language model to generate complete plans in one LLMs reasoning steps. The accuracy of this method is limited with the 'next token prediction' paradigms~\cite{bubeck2023sparks}.

\subsubsection{Tree-of-Thought(ToT)} The Tree-of-Thought models the generation of a sequence of plans for a search task~\cite{yao2023tree}.  Specifically, each node in the tree represents an atomic plan.  The goal of ToT is to find an optimal plan trajectory that can solve a given question.  The ToT typically consists of two main components:
1. Generator, which responds to generate alternative atomic plans at each node of the tree. 2. Evaluators, which are charged with evaluating generated plans to judge whether they meet set rules. In this study, we design these two components for multi-hop visual reasoning tasks as shown in the next section.
For all ToT based methods, we employ a depth-first search algorithm as the search strategy.
\subsection{Tree-of-Mixed-Thought(ToMT)}
\label{sec:tomt}
We aim to utilize a one-stop approach to minimize the number of ToT searches and thereby maximize efficiency. To achieve this objective, we design two strategies for their integration.

\subsubsection{ToT-One-Stop(ToT-OS)}
 
ToT-One-Stop is a greedy strategy centered around ToT as shown in Algorithm ~\ref{alg:tot-dfs}. When ToT identifies a suitable plan, it utilizes a one-stop generation approach to generate all the remaining plans. The search procedure is considered complete if the generated set of remaining plans adheres to the predefined rules. If the generated set fails to fulfill the predefined rules, the ToT continues its search for the next individual plan. This iterative process persists until the max step is reached or the stop\_sign is detected.  Since ToT adopts a depth-first search strategy, it always traverses the left node first. In this way, this approach can be easily implemented by simply equipping the left node with a one-stop generation function.

\subsubsection{ToT-Block}
The ToT-Block approach presents a more intuitive integration of ToT and one-stop approaches as shown in Algorithm~\ref{alg:tot-dfs} . In the ToT-Block strategy, instead of generating a single-step plan, each node generates a multi-step plan, thereby significantly reducing the search path to generate the entire plan trajectory. For example, for a 4-hop question, ToT needs at least 4 iteration steps to generate the whole plan, while the ToT-Block under 2 block size situation (each time generate the next 2 plans), may need only 2 iteration steps to generate the whole plans. (if $n>$ \#hop, it's equal to multi-iteration for one-stop)

\subsubsection{Generator}
\label{sec:generator}
Each node generation is accompanied by a predetermined prompt that concatenates the already generated steps to serve as guidance for generating the next step or the remaining steps. Additionally, in order to enhance the diversity of generated outputs and enable the tree search to explore different search paths more effectively, we incorporate the random selection of different examples as demonstrations during each generation process. In the original ToT method, a single-step atomic plan is generated each time. In the ToT-One-Stop approach, the generator directly produces all remaining plans at the left node when the conditions are met. In contrast, the ToT-Block generator produces the next $k$ plan sequence each time.

\begin{algorithm}
    \caption{ToMT-DFS(s, G, V, d, b, k, T, sn)}
    \begin{algorithmic}[1]\label{alg:tot-dfs}
        \STATE Plan generator $G()$, state evaluator $V()$, max\_step $T$, tree branch size $b$, start\_one\_stop sn, current\_depth $d=1$, start\_prompt $s$, block size $k$, stop\_sign = 'ans'
        \STATE $DFS(s)$
        \STATE $t = t + 1$
        \IF{$t> T$}
            \STATE return True
        \ENDIF
        \FOR {t=$1,2,\ldots,b$}
            \STATE $s^{'} = G(p_{\theta}, s,d,i,k)$
            \IF{$V(s^{'})$ is True and stop\_sign in $s^{'}$}
                \STATE return True
            \ENDIF
            \IF{$V(s^{'})$ is right}
                \STATE $ d = d + 1$
                \STATE $s^{'} = s + s^{'}$ \small \#concat the generate content and prompt
                \STATE $DFS(s^{'})$
                \STATE $d = d - 1$
            \ENDIF
        \ENDFOR
    \end{algorithmic}
    \begin{algorithmic}[1] 
        \STATE $G()$ for ToT-One-Stop 
        \IF{$d>=sn$ and i=0}
            \STATE generate all remains plans
        \ELSE
            \STATE generate next single plan
        \ENDIF
    \end{algorithmic}
    \begin{algorithmic}[1]
        \STATE $G()$ for ToT-Block
            \STATE generate k plans
    \end{algorithmic}
\end{algorithm}
\subsubsection{Evaluator} 
 
Evaluating generated plans is a crucial issue that directly impacts the efficiency of tree search. However, in long-range planning problems, the accuracy of individual steps cannot ensure overall accuracy, as the solution often depends on the accuracy of the entire sequence. Therefore, evaluating the accuracy of individual steps poses a significant challenge. In this study, we observed that code-like plans generated by LLM are likely to be effective for the overall structure if the usage of functions and parameters is correct. Thus, we employ two approaches to assess the generated plans: 1) checking whether the generated code can run correctly, including the use of appropriate functions and parameters, and 2) interacting with the scene graph using in certain tools, such as query\_part, to verify whether the parameters are consistent with the elements in the current scene graph.

It should be noted that even if the generated code meets the above criteria, it cannot guarantee the correctness of the final result, as individual steps may be correct but the overall logic may still be flawed. We refer to plans that meet our evaluation criteria as accurately evaluated plans, and the accuracy of plans that yield correct answers as the gold standard. We record the inconsistency between these two measures in  \Cref{tab:ablation_backtrack_rate}.
\section{Experiments and discussion}

\subsection{Experiment Setup}
In all experiments, we utilized ChatGPT (chat-gpt-tubor-0631)~\cite{openai2023chat} with a temperature setting of 1 to ensure a diverse generation. Each experiment was run three times, and the average results were reported.

For all experiments, we employed a consistent prompt structure inspired by \cite{vemprala2023chatgpt}. The prompt structures are organized to the following template: 

\begin{verbatim}
{instruction}
{tool description}
{examples}
{question}
Answer:
\end{verbatim}

Further details of the prompt can be found in the Appendix. 

Regarding the depth-first search, we imposed a maximum step limit of 30, as we observed that the majority of problems could be solved within a search space of 30 steps~\footnote{
The limit on the maximum searching step has greater effects on the standard ToT algorithm than on ToT-OneStop and ToT-Block. For details please refer to the supplementary material.  
}. Additionally, for the ToT and ToT-OS experiments, we set the tree branching factor to 3. For the ToT-Block experiment, we increased the branching factor to 5, as the inclusion of blocks reduces the depth of exploration and necessitates a wider search space.
\begin{table*}[!h]
    \centering
    \setlength{\tabcolsep}{1.1mm}
    \begin{tabular}{clc@{\hspace*{6mm}}c@{\hspace*{5mm}}c@{\hspace*{6mm}}c@{\hspace*{8mm}}c}
    \toprule
    & & & \multicolumn{4}{c}{{\it LLM-based planning algorithms}}\\
    \cmidrule{4-7}
    & & avg. \#hop & {\bf One-stop} & {\bf ToT} & {\bf ToT-OS} & {\bf ToT-Block} \\
    \midrule
    \multirow{2}{*}{
        \rotatebox[origin=c]{0}{{\it Sequence}}
    } & Short Rel & 4.1 & 39.33 & 52.66 (5.48)  & 56.66 (3.04) & 59.33 (3.30)\\
      & Long Rel  & 8.3 &  7.66 & 56.00 (15.90) & 53.00 (9.63) & 55.00 (10.49)\\
    \midrule
    \multirow{3}{*}{
        \rotatebox[origin=c]{0}{{\it Parallel}}
    } & Sum     & 5.6 & 86.00 & 98.66 (6.39)  & 100.00 (2.47) & 93.44 (3.22)\\
      & Compare & 5.9 & 52.00 & 59.66 (6.91)  &  63.00 (2.39) & 59.00 (3.89)\\
      & Logic   & 7.2 & 31.00 & 65.00 (12.81) &  69.00 (5.50) & 70.66 (7.67)\\
    \midrule
    \multirow{3}{*}{
        \rotatebox[origin=c]{0}{{\it Backtrack}}
    } & Query Part & 5.5 & 74.00 & 77.00 (5.06) & 78.66 (2.08) & 78.33 (3.12)\\
      & Exist      & 5.6 & 53.00 & 82.00 (7.58) & 90.33 (4.40) & 86.33 (5.25)\\
      & Count      & 5.9 & 72.66 & 85.66 (8.25) & 90.00 (3.77) & 92.00 (5.13)\\
    \midrule
    \makecell{
        {\it Multi} \\ {\it Backtrack}
    } & Analogy & 6.7 & 70.33 & 89.66 (8.57) & 93.00 (3.12) & 92.33 (4.37)\\ 
         \bottomrule
    \end{tabular}
    \caption{Main Result. We report the mean accuracy for all method. We include the average step consumption of tree search methods within the brackets.}
    \label{tab:main_res}
\end{table*}
\subsection{Main Results}

\Cref{tab:main_res} summarizes the performance of the competing algorithms on different types of questions. 
   In the comparison, we include the One-stop and the ToT as two extreme baselines, standing for the solely intuition-based thinking style (fast) and the most elaborated thinking style (slow).
The results are grouped by the topology of the question's logic structures, depending on whether there include branches or back-tracking pathways.

\begin{table}[!ht]
    \centering
    \setlength{\tabcolsep}{1.1mm}
    \begin{tabular}{clcccccccccccccc}
    \toprule
    & {} & {ToT-OS-ratio} & {ToT-Block-ratio}\\
    \midrule
    \multirow{2}{*}{
        \rotatebox[origin=c]{0}{{\it Sequence}}
    } & Short Rel &  1.78 &  1.66  \\
      & Long Rel  & 1.71  &  1.51  \\
    \midrule
    \multirow{3}{*}{
        \rotatebox[origin=c]{0}{{\it Parallel}}
    } & Sum & 2.64 &  1.91 \\
      & Compare & 2.89  & 1.90 \\
      & Logic    & 2.32 & 1.67 \\
    \midrule
    \multirow{3}{*}{
        \rotatebox[origin=c]{0}{{\it Backtrack}}
    } & Query Part &  2.43 & 1.62 \\
      & Exist      & 1.72 & 1.44 \\
      & Count      & 2.18 & 1.60 \\
    \midrule
    \multirow{1}{*}{
    } 
    \makecell{
        {\it Multi} \\ {\it Backtrack}
    } & Analogy & 2.74  & 1.96 \\ 
         \bottomrule
    \end{tabular}
    \caption{Result for reasoning step saving index.}
    \label{tab:ablation_backtrack_rate}
\end{table}

The primary conclusion is that the ToT-OS algorithm is both accurate and efficient.
Compared to the ToT algorithm, ToT-OS achieves the same level of plan accuracy while saving more than half the number of reasoning steps across all types of questions. 
In fact, ToT-OS slightly improves the planning accuracy in all situations except for the Long relation case.
The ability to perform complex reasoning is inherited from the tree-guided searching framework. 
We find that all three tree-based searching algorithms significantly outperform the naive one-stop reasoning by a large margin. 
However, the standard ToT algorithm is very time-consuming. For hard tasks such as Long relation and Logic, the average reasoning steps are 17.24 and 12.81, respectively.
Benefiting from the fast one-stop thinking, ToT-OS only spends 9.63 and 5.50 average reasoning steps in the above two cases.
To better capture the efficiency of our proposed methods, we define the reasoning step saving index (RSSI) as the ratio between the reasoning steps of ToT and the ones of ToT-OS  (or ToT-Block). 
\Cref{tab:ablation_backtrack_rate} shows the index for all types of questions. The index of ToT-OS is consistently higher than 1.7 and reaches its maximum at the Compare case (2.89).  The rate of speedup in reasoning is more evident for relatively simple questions.
The balance between performance and resource consumption makes ToT-OS a desirable alternative to the standard ToT algorithm in practical deployment.

While ToT-OS relies on a greedy use of fast thinking strategy at each tree node, the ToT-Block algorithm is more structured and conservative.
At each node, ToT-Block adopts a fix-sized fast-thinking block.
In this way, it resembles standard tree-based searching while improving efficiency by generating multiple plan steps at a time.
From~\Cref{tab:main_res}, the performance of ToT-Block is similar to the ones of ToT and ToT-OS. While being more efficient than the standard ToT, its reasoning step-saving index is generally small than the one of ToT-OS.
It is worth noting that ToT-Block is particularly suitable for processing questions of the sequence reasoning type, for which
ToT-Block has the same level of plan accuracy as the standard ToT and considerably reduces the reasoning steps with RSSIs larger than 1.5.
These results indicate that ToT-Block can effectively exploit the tree-based searching for hard questions and improve reasoning efficiency.

\subsection{Analysis}

\begin{table}[!ht]
    \centering
    \setlength{\tabcolsep}{1.1mm}
    \begin{tabular}{ccccc}
    \toprule
    {}   & One-Stop & ToT & ToT-OS& ToT-Block \\
    \midrule
    Acc.  & -0.500 & -0.095 & -0.206 & -0.176 \\
    \#Step & - & 0.912 & -0.005 & 0.857 \\
    NoBack & - &  -0.54  & -0.46  & -0.557 \\
    \bottomrule
    \end{tabular}
    \caption{The correlation between hop with ACC, inference step, and NoBack number.}
    \label{tab:correlation_acc_step_inference_step}
\end{table}

\begin{table*}[!ht]
    \centering
    \setlength{\tabcolsep}{1.1mm}
    \begin{tabular}{clcccccccccccccc}
    \toprule
    & & \multicolumn{2}{c}{\bf ToT} & & \multicolumn{2}{c}{\bf ToT-OS} & & \multicolumn{2}{c}{\bf ToT-block}  \\
     \cmidrule{3-4}\cmidrule{6-7} \cmidrule{9-10}
    & {} & {NoBack} & {Incons} & & {NoBack} & {Incons} & & {UnBack} & {Incons} \\
    \midrule
    \multirow{2}{*}{
        \rotatebox[origin=c]{0}{{\it Sequence}}
    } & Short Rel & 35.00 & 38.33  & & 38.33 & 32.66 & & 44.33 & 29.33  \\
      & Long Rel  &  8.00 & 14.33  & & 10.66 & 3.00  & & 8.00 & 3.00  \\
    \midrule
    \multirow{3}{*}{
        \rotatebox[origin=c]{0}{{\it Parallel}}
    } & Sum     & 81.33 & 1.00  & & 87.00 & 0 & & 83.33 & 0 \\
      & Compare & 50.66 & 38.33  & & 55.00 & 36.33 & & 53.00 & 37.00 \\
      & Logic   & 15.66 & 3.60  & & 32.00 & 3.66 & & 28.00 & 2.33 \\
    \midrule
    \multirow{3}{*}{
        \rotatebox[origin=c]{0}{{\it Backtrack}}
    } & Query Part & 72.00 & 0.33  & & 73.66 & 0.66 & & 71.66 & 0 \\
      & Exist      & 53.00 & 10.33  & & 54.66 & 2.33 & & 50.33 & 2.33 \\
      & Count      & 65.33 & 1.00  & & 72.33 & 1.0 & & 63.66 & 0.66 \\
    \midrule
    \multirow{1}{*}{
    } 
    \makecell{
        {\it Multi} \\ {\it Backtrack}
    } & Analogy & 58.66 & 6.00 & & 73.66 & 4.33 & & 71.33 & 3.33 \\ 
         \bottomrule
    \end{tabular}
    \caption{The number of Not backtrack and inconsitency.}
    \label{tab:ablation_backtrack_rate}
\end{table*}

\begin{table*}[!ht]
    \centering
    \setlength{\tabcolsep}{1.1mm}
    \begin{tabular}{clccccccccccc}
    \toprule
    & & \multicolumn{3}{c}{\bf ToT-OS (start depth)} & & \multicolumn{3}{c}{\bf ToT-block (k-block)}  \\
    \cmidrule{3-5} \cmidrule{7-9}
    & {} & 1 & 2 & 3 & &  2 & 3 & 4\\
    \midrule
    \multirow{2}{*}{
        \rotatebox[origin=c]{0}{{\it Sequence}}
    } & Short Rel & 59.41(1.80) & 58.33(3.17)  & 58.00(4.56) & & 59.33(3.30) & 60.33(2.59) & 58.00(1.96) \\
      & Long Rel  &  48.33(10.89) & 53.00(9.26) & 48.00(12.85) & & 55.00(10.49) & 48.00(7.28) & 48.00(5.87) \\
    \midrule
    \multirow{3}{*}{
        \rotatebox[origin=c]{0}{{\it Parallel}}
    } & Sum     & 99.33(1.47) & 100.00(2.45) & 99.33(3.43) & & 100.00(3.39) & 99.66(2.50) & 98.66(2.21) \\
      & Compare & 62.66(1.59) & 63.00(2.39) & 62.00(4.14) & & 61.33(3.63) & 61.33(2.80) & 61.33(2.52) \\
      & Logic   & 67.33(4.36) & 69.00(5.50) &  67.66(6.19) & & 70.66(7.67) & 69.33(5.14) & 58.00(4.47) \\
    \midrule
    \multirow{3}{*}{
        \rotatebox[origin=c]{0}{{\it Backtrack}}
    } & Query Part & 78.00(1.11) & 78.66(2.08) & 79.00(3.07) & & 78.33(3.12) & 78.33(2.08) & 79.33(2.08) \\
      & Exist      & 90.00(3.33) & 90.33(4.40) & 90.66(5.81) & & 86.66(5.25) & 83.33(4.00) & 81.00(2.63) \\
      & Count & 90.00(2.33) & 90.00(3.77) & 90.66(5.06) & & 92.00(5.13) & 90.00(2.87) & 86.66(2.64) \\
    \midrule
    \multirow{1}{*}{
    } 
    \makecell{
        {\it Multi} \\ {\it Backtrack}
    } & Analogy & 91.66(2.08) & 93.00(3.12) & 90.66(4.20) & & 92.33(4.37) & 90.66(3.02) & 90.66(2.49) \\ 
         \bottomrule
    \end{tabular}
    \caption{Ablation Study. The accuracy of ToT-OS under different start depth and ToT-block under different block settings.}
    \label{tab:ablation_block_size}
\end{table*}

\subsubsection{Question Types\label{testcitesec}} 
Next, we investigate how question types affect the algorithm's performance. 
We find that all four algorithms share a similar overall tendency of performance across different types of questions and the Sequence reasoning type has the lowest performance.
This may be attributed to the deficiency of LLMs when doing chained reasoning where each step depends on the answer of its immediate predecessor. As the length of the chain grows, LLMs start to lose track even with the help of tree-based searching.
Nonetheless, the topology of the question's logic structure generally does not decide the plan's accuracy. 
Within a type, there could be a large variation in plan accuracy. 
The Parallel type exemplifies this situation where ToT-OS never makes a mistake on the Sum type and only has an accuracy of 63\% on the Compare case.
We will dive deeper into studying why our algorithms perform poorly on some individual question types such as Logic and Query Part in the later discussion.

\subsubsection{Hop Number}
The question type only partially reflects the structure of the underlying questions.
Another important metric is the number of logically dependent elements in the question.
We refer to this number as the hop number of a question.
The hop number decides the minimum number of inference steps needed to answer a question and, thus, measures the complexity of this question.
In the third column of~\Cref{tab:main_res}, we list the average hop numbers of different types of questions.
To show how the hop number relates to the algorithm's performance, we further compute the correlation between the hop number and the planning accuracy and the reasoning step, 
respectively. 
The results are summarized in ~\Cref{tab:correlation_acc_step_inference_step}.
Interestingly, when introducing fast thinking the plan accuracy shows a negative correlation with the hop number. This is most evident for the One-Stop case where fast thinking is the sole strategy. The results make sense as the fast thinking speeds up the reasoning process by providing shortcuts to elaborate tree-based searching.
Next, we note a very strong positive correlation between the number of reasoning steps and the hop number for the ToT and the ToT-Block algorithms.
On the opposite, the number of reasoning steps for ToT-OS barely correlates to the hop number. 
We attribute this contrasting behavior to the fact that the ToT and the ToT-Block largely retain the structure of tree-based searching while the greedy application of One-Stop fast thinking in ToT-OS breaks this structure.

\subsubsection{Spatial Relation}
It is very interesting to note that the reasoning on spatial relation post a serious challenge to the LLM-based planning algorithms under our consideration.
Even though equipped with a powerful tree-based searching strategy, all the LLM-based planning algorithms perform poorly for the questions of the Short Rel, the Long Rel, and the Logic types. 
These types of questions in our dataset are characterized by requiring long reasoning sequences to infer the spatial relation among objects. 
In fact, only those three types of questions contain words, ``right'', ``left'', ``behind'' and ``front'', that directly describe the spatial relation.
Two sources might account for the poor performance (i.e., low plan accuracy and large reasoning steps) in these three cases.
First, the LLMs have difficulties to recognize and understand spatial relations;
Second, current LLMs have a deficiency of handling a long chain of inference. 

\subsubsection{Semantic Issue}
For the Query Part and Compare questions, the LLM-based planning algorithms also perform poorly.
A closer look reveals that for the Query Part type, the algorithms often fail to identify compositional parts. For example, the part 'central support' and 'leg bar' are often recognized to 'support' or 'leg' by LLM.
This kind of error is fact-oriented and can be fixed by telling the LLM knowledge about the objects relevant to the question (Simply add the sentence "The part include 'central support',' leg bar'" to the prompt can reduce this type of error). 
The deficiency in the Compare questions has a slightly different nature. In this case, the LLM tends to mistake the branches used for comparison.  For example, give the question ``How many more legs are in the chair with purple arm horizontal bars, than the number of wheels in the cart?'', the LLM may count the number of legs in the cart. Such errors are more logic-oriented rather than fact-oriented.
To address the fact-oriented deficiency, one may integrate the LLM with an external knowledge base that stores common sense facts. 
The logic-oriented deficiency could be also alleviated by combining the LLM with other logic-aware processing systems such as Mathematica~\cite{wolfram}.
However, a complete solution may require a substantial change of the current 'next token prediction' paradigm of LLMs~\cite{bubeck2023sparks}.

\subsubsection{Backtracking and Consistency}
The ability to backtrack upon mistake lies at the core of tree-based searching algorithms. 
The need for back-tracking indicates that the underlying question is hard to answer.
One naturally expects that those hard questions would require sophisticated search algorithms to solve.
In~\Cref{tab:ablation_backtrack_rate}, we show the average number of correctly answered questions without back-tracking for each type of question. 
The tendency is clear that for simple questions such as those of the Sum type, a large number of questions can be resolved without back-tracking, while for hard questions such as those of the Long Rel type, back-tracking is needed for almost all questions.
We find, indeed, for questions of the Long Rel and Logic types, a simple One-Stop algorithm achieves very low plan accuracy.
Moreover, we compute the correlation between the number of correct plans without back-tracking and the number of hops in different types of questions. The results are shown in~\Cref{tab:correlation_acc_step_inference_step} where we observe significant negative correlations. 
This confirms our intuition that hard questions (with large hop numbers) rely more on the back-tracking mechanism for generating correct plans.

Recall that in this work the back-tracking in searching algorithms is triggered by a sanity check. The ultimate goal of the back-tracking search is to guarantee an executable plan.
It is important to investigate the proportion of correct plans among those executable ones. This value conveys the robustness and consistency of the underlying searching algorithm. \Cref{tab:ablation_backtrack_rate} illustrates the inconsistency number of each algorithm for each type of question, which is defined as the average number of incorrect plans. 
We note that one should not confuse the inconsistent number and the hardness of questions.
As an example, the Logic type is much harder than the Sum type, in the sense that all algorithms have lower plan accuracy and lower NoBack number for the former case. 
Nonetheless, these two types have comparable inconsistent numbers.

\subsection{Ablation Study}

\subsubsection{When to start one-stop search.}
In the main results, we chose the start depth to be 2 for the ToT-OS algorithm.
Namely, the fast One-Stop thinking only applies to nodes at depth $d > 1$. 
The starting point can affect the performance of the algorithm.
If fast thinking operates from the beginning, then the probability will become higher for the incorrect plans to pass the sanity check, resulting in degraded performance.
If one only applies the One-Stop to deep nodes, then the algorithm can hardly benefit from the fast thinking.
Columns 3 to 5 of~\Cref{tab:ablation_block_size} display the performances of ToT-OS for different start depths.
It is easy to see that small or large start depth will lead to sub-optimal results and with $d = 2$ ToT-OS reaches its best performance.

\subsubsection{Block size}
In~\Cref{tab:ablation_block_size} we also study the influence of block size on the ToT-Block algorithm (columns 6 to 8). 
We find that increasing block size consistently leads to fewer reasoning steps.
However, the plan accuracy generally gets worse for larger block sizes. 
This reflects the trade-off between reasoning efficiency and the ability to perform tree-based elaborated searching.
\section{Conclusion}
In this study, we propose two combined strategies by integrating the one-stop generation and ToT generation methods, which significantly reduce the search steps and resource consumption while preserving the ToT's backtracking and iterative generation features. Our approach outperforms ToT in accuracy and inference steps for most test problems. However, our method shares the same limitations as ToT, including the need for an efficient evaluation method to determine whether the current search node meets the requirements, which is challenging to achieve in most scenarios. Additionally, we chose a dataset that can fully express visual information using a scene graph to helps independently evaluate the LLM's accuracy in multi-hop visual reasoning tasks, ensuring the tool's accuracy. Nevertheless, in real-world situations, expert modules often have uncertain accuracy, making it more challenging for LLM to find effective planning. 
Notwithstanding these constraints, we posit that our methodology offers a novel perspective for investigating the proficient use of LLM-generated, code-like planning. Furthermore, we anticipate broadening the application of this approach to encompass a wider range of scenarios in forthcoming studies.

\nocite{langley00}
\bibliography{agm}
\bibliographystyle{icml2023}

\newpage
\appendix
\onecolumn

\section{Appendix}
\subsection{Dataset Details} 
We curated seven question types from the PTR dataset and two question types from the CLEVR dataset (Logic Both, Long Spatial Relation). These questions have been grouped into four distinct categories based on their respective reasoning structures. To illustrate each of these question types, we present the following examples.

\subsubsection{Sequence}
\begin{verbatim}
1. Short Sptatial Relation 

Question: how many things are behind the table with six red drawers?
Answer:
Step 1:obj1 = filter_object("table",all_obj)
Step 2:obj2 = filter_part(["six","red","drawer"],obj1)
Step 3:obj3 = query_relation("behind",obj2)
Step 4:ans = count_object(obj3)

2. Long Spatial Relation

Question: What size is the yellow ball behind the sphere that is on the right 
side of the object that is behind the tiny yellow matte thing?
Answer:
Step 1:obj1 = filter_part(["tiny","yellow","matte"],all_obj)
Step 2:obj2 = query_relation("behind",obj1)
Step 3:obj3 = query_relation("right",obj2)
Step 4:obj4 = filter_object("sphere",obj3)
Step 5:obj5 = query_relation("behind",obj4)
Step 6:obj6 = filter_object("ball",obj5)
Step 7:obj7 = filter_part(["yellow"],obj6)
Step 8:ans = query_size(obj7)
\end{verbatim}

\subsubsection{Parallel}
\begin{verbatim}
1. Sum 

Question: what is the sum of the number of drawers in the table with cyan legs, 
and the number of doors in the table with two blue legs?
Answer:
Step 1:obj1 = filter_object("table",all_obj)
Step 2:obj2 = filter_part(["cyan", "leg"],obj1)
Step 3:num1 = count_part("drawer",obj2)
Step 4:obj3 = filter_object("table",all_obj)
Step 5:obj4 = filter_part(["two","blue","leg"],obj3)
Step 6:num2 = count_part("door",obj3)
Step 7:ans = sum(num1,num2)

2. Compare

Question: are there an equal number of legs in the object with three red leg
bars, and wheels in the cart with a cyan body?
Answer:
Step 1:obj1 = filter_object("cart",all_obj)
Step 2:obj2 = filter_part(["one","cyan","body"],obj1)
Step 3:num1 = count_part("wheel",obj2)
Step 4:obj3 = filter_part(["red","leg bar"],all_obj)
Step 5:num2 = count_part("leg",obj3)
Step 6:ans = equal(num1,num2)

3. Logic Both

Question: The metal object that is both on the right side of the small red
thing and in front of the large green metal thing is what color?
Answer:
Step 1:obj1 = filter_part(["small","red"],all_obj)
Step 2:obj2 = query_relation("right",obj1)
Step 3:obj3 = filter_part(['large',"green","metal"])
Step 4:obj4 = query_relation("front",obj3)
Step 5:obj5 = intersection(obj2,obj4)
Step 6:obj6 = filter_part(["metal"],obj5)
Step 7:ans = query_color(obj6)
\end{verbatim}

\subsubsection{Backtrack}
\begin{verbatim}
1. Query Part 

Question: what is the category of the brown part of the thing that is of the
same category as the thing with blue legs?
Answer:
Step 1:obj1 = filter_part(["blue","leg"],all_obj)
Step 2:category1 = query_category(obj1)
Step 3:obj2 = filter_category(category1,all_obj)
Step 4:obj3 = exclude_object(obj1,obj2)
Step 5:ans= query_part("brown",obj3)

2. Exist

Question: are there any objects that have the same number of leg bars as the
chair with one red back?
Answer:
Step 1:obj1 = filter_object("chair",all_obj)
Step 2:obj2 = filter_part(["one","red","back"],obj1)
Step 3:num1 = count_part("leg bar",obj2)
Step 4:obj3 = filter_part([num1,"leg bar"],all_obj)
Step 5:obj4 = exclude_object(obj2,obj3)  
Step 6:ans = exist(obj4)

3. Count

Question: what is the number of the legs of the thing that has the same color
of seat as the chair with three leg bars?
Answer:
Step 1:obj1 = filter_object("chair",all_obj)
Step 2:obj2 = filter_part(["three","leg bar"],obj1)
Step 3:color1 = query_color("seat",obj2)
Step 4:obj3 = filter_part([color1,"seat"],all_obj)
Step 5:obj4 = exclude_object(obj2,obj3)  
Step 6:ans = count_part("leg",obj4)
\end{verbatim}

\subsubsection{Multi-Backtrack}
\begin{verbatim}
1. Analogy

Question: the bed with a purple sleep area has certain positional relation to
the bed with one green sleep area. by analogy, how many objects does the object
with eight leg bars have the same positional relation to?
Answer:
Step 1:obj1 = filter_object("bed",all_obj)
Step 2:obj2 = filter_part(["one","purple","sleep area"],obj1)
Step 3:obj3 = filter_object("bed",all_obj)
Step 4:obj4 = filter_part(["one","green","sleep area"],obj3)
Step 5:relation1 = get_relation(obj2,obj4)
Step 6:obj5 = filter_part(["eight","leg bar"],all_obj)
Step 7:obj6= filter_relation(relation1,obj5,all_obj)
Step 8:ans = count_object(obj6)
\end{verbatim}

\subsection{Scene Graph Example}
This section presents examples of scene graphs obtained from the PTR dataset and CLEVR dataset. It is worth note that, in the PTR dataset, the numerical value assigned to an object signifies its corresponding part number. In the case of the CLEVR dataset, the scene graphs solely encompass three attributes, namely color, size, and materials.
\subsubsection{PTR Dataset}
\begin{verbatim}
{"question": "the chair with four brown arm vertical bars has certain
positional relation to the chair with gray legs. by analogy, the thing 
with three blue legs has the same positional relation to an object of 
what category?"
"answer": "Chair",  
"relationships": 
{"above": [[], [], []], 
"behind": [[], [0], [0, 1]], 
"below": [[], [], []], 
"front": [[1, 2], [2], []], 
"left": [[1, 2], [], [1]], 
"right": [[], [0, 2], [0]]}, 
"objects": [
{"Chair0": {"back": ["cyan", 1], "leg": ["gray", 4], "seat": ["gray", 1]}},
{"Table0": {"door": ["purple", 1], "leg": ["blue", 3], "top": ["yellow", 1]}},
{"Chair1": {"arm horizontal bar": ["green", 2], "arm vertical bar": ["brown",
4], "back": ["brown", 1], "leg": ["cyan", 4], "leg bar": ["green", 2], "seat":
["brown", 1]}}]}
\end{verbatim}

\subsubsection{CLEVR Dataset} 
\begin{verbatim}
{"question": "The large thing that is both on the left side of the purple shiny
object and behind the tiny gray metallic ball is what color?", 
"answer": "brown",  
"relationships": 
{"right": [[1, 2, 4, 7], [2, 7], [], [0, 1, 2, 4, 7], [1, 2, 7], [0, 1, 2, 3,
4, 6, 7], [0, 1, 2, 3, 4, 7], [2]], 
"behind": [[1, 2, 3, 4, 5, 6, 7], [4, 7], [1, 4, 7], [1, 2, 4, 7], [], [1, 2,
3, 4, 7], [1, 2, 3, 4, 5, 7], [4]], 
"front": [[], [0, 2, 3, 5, 6], [0, 3, 5, 6], [0, 5, 6], [0, 1, 2, 3, 5, 6, 7],
[0, 6], [0], [0, 1, 2, 3, 5, 6]], 
"left": [[3, 5, 6], [0, 3, 4, 5, 6], [0, 1, 3, 4, 5, 6, 7], [5, 6], [0, 3, 5,
6], [], [5], [0, 1, 3, 4, 5, 6]]}, 
"objects": 
[{"cube0": ["gray", "large", "rubber"]}, 
{"sphere0": ["brown","small", "metal"]}, 
{"cube1": ["blue", "large", "rubber"]}, 
{"cylinder0": ["brown", "large", "rubber"]}, 
{"cube2": ["purple", "small", "metal"]}, 
{"sphere1": ["gray", "small", "metal"]}, 
{"cube3": ["gray", "small", "rubber"]}, 
{"cube4": ["blue", "small", "metal"]}]}
\end{verbatim}

\subsection{Tools Details}
We developed various tools to help support answering this question. Certain tools are designed to facilitate interaction with scene graphs, while others serve specific functions such as sum, comparison (e.g. fewer\_than), and counting. We categorize these tools into four types, namely Filter, Algebra, Relation and Attribution. Here we provide a detailed introduction to these tools. 

\subsubsection{Filter}
\begin{verbatim}
filter_object(category, object_list): Take a category and a list of objects, and
return the objects with the category.

filter_part(part_list, object_list): Take a list of parts or attributes of an
object and an object list,and return the object with the attribute or part.

get_relation(object,object): Take two objects, and return the relation between
them.

exclude_object(object,object_list): Take an object and an object list, and 
return the object list excluding the input object.

filter_category(category,object_list): Take an category and an object list, and
return the objects with the input category.
\end{verbatim}

\subsubsection{Algebra}
\begin{verbatim}
count_part(part,object): Take a part and an object, and return the number of
parts in the object.

count_object(object_list): Take an object list, and return the number of objects
in the list.

sum(num1,num2): Take two numbers, and return the sum of the two numbers.

equal(input1, input2): Take two objects or numbers or attributes, and return a
boolean value based on the comparison.

more_than(num1,num2): Take two numbers, and return if the first number more than
second number.

few_than(num1,num2): Take two numbers, and return if the first number less than
second number.

intersection(object_list,object_list): Take two lists of objects, and return the
intersection terms.
\end{verbatim}

\subsubsection{Relation}
\begin{verbatim}
query_relation(relation,object): Take a relation and an object, and return the
thing with the relation to the object.

filter_relation(relation,thing,object_list):  Take a relation, an object, and a
list of objects, and return the objects in the list of objects that has the 
relation to the input object.

get_relation(object,object): Take two objects, and return the relation between 
them.
\end{verbatim}

\subsubsection{Attribution}
\begin{verbatim}
query_category(object_or_part): Take a part or an object, and return the 
category of the part or object.

query_color(part, object): Take a part and an object, and return the color of 
the part in the object.

query_part(part,obj): Take an attribute and an object, and return the category 
of the part in the object.

query_size(object): Take an object, and return the size of the object.
\end{verbatim}

\subsection{Prompt Details}
The structure of the prompt consists of four parts:  instruction, tool description, example, and  question. For all experiments, we use the same prompt structure the number of examples is set to four.
The prompt template is as follows:

\textbf{Prompt Template}
\begin{verbatim}
{instruction}
{tool description}
{example}
{question}
\end{verbatim}

\textbf{Prompt Example}
Here is a whole prompt with the ``sum'' questions as an example.
\begin{verbatim}
Output python code with the sequence of steps to solve the given question. 

filter_part(part_list, object_list): Take a list of parts or attributes of an 
object and an object list, and return the object with the attribute or part.
get_relation(object,object): Take two objects, and return the relation between 
them.
filter_relation(relation,thing,object_list):  Take a relation, an object, and a
list of objects, and return the objects in the list of objects that has the 
relation to the input object.
count_object(object_list): Take an object list, and return the number of objects
in the list.
exist(object_list): Take an object list, and return a boolean value to stand by
if the object list is null.
query_category(object_or_part): Take a part or an object, and return the 
category of the part or object.

Question: the bed with a purple sleep area has certain positional relation to
the bed with one green sleep area. by analogy, how many objects does the object
with eight leg bars have the same positional relation to?
Answer:
Step 1:obj1 = filter_object("bed",all_obj)
Step 2:obj2 = filter_part(["one","purple","sleep area"],obj1)
Step 3:obj3 = filter_object("bed",all_obj)
Step 4:obj4 = filter_part(["one","green","sleep area"],obj3)
Step 5:relation1 = get_relation(obj2,obj4)
Step 6:obj5 = filter_part(["eight","leg bar"],all_obj)
Step 7:obj6= filter_relation(relation1,obj5,all_obj)
Step 8:ans = count_object(obj6)

Question: the table with purple top has certain positional relation to the table
with doors. by analogy, is there an object that cart with three wheels has the 
same positional relation to?
Answer:
Step 1:obj1 = filter_object("table",all_obj)
Step 2:obj2 = filter_part(["purple","top"],obj1)
Step 3:obj3 = filter_object("table",all_obj)
Step 4:obj4 = filter_part(["door"],obj3)
Step 5:relation1 = get_relation(obj2,obj4)
Step 6:obj5 = filter_object("cart",all_obj)
Step 7:obj6 = filter_part(["three","wheel"],obj5)
Step 8:obj7= filter_relation(relation1,obj6,all_obj)
Step 9:ans = exist(obj7)

Question: the thing with two green leg bars has certain positional relation to 
the object with four brown legs. by analogy, the chair with gray legs has the 
same positional relation to an object of what category?
Answer:
Step 1:obj1 = filter_part(["two","green","leg bar"],all_obj)
Step 2:obj2 = filter_part(["four","brown","leg"],all_obj)
Step 3:relation1 = get_relation(obj1,obj2)
Step 4:obj3 = filter_object("chair",all_obj)
Step 5:obj4 = filter_part(["gray","leg"],obj3)
Step 6:obj5= filter_relation(relation1,obj4,all_obj)
Step 7:ans = query_category(obj5)

Question: the chair with arm horizontal bars has certain positional relation to
the chair with blue seat. by analogy, is there an object that chair with red 
back has the same positional relation to?
Answer:
Step 1:obj1 = filter_object("chair",all_obj)
Step 2:obj2 = filter_part(["arm horizontal bar"],obj1)
Step 3:obj3 = filter_object("chair",all_obj)
Step 4:obj4 = filter_part(["blue","seat"],obj3)
Step 5:relation1 = get_relation(obj2,obj4)
Step 6:obj5 = filter_object("chair",all_obj)
Step 7:obj6 = filter_part(["red","back"],obj5)
Step 8:obj7= filter_relation(relation1,obj6,all_obj)
Step 9:ans = exist(obj7)

Question: the table with one yellow top has certain positional relation to the 
bed with a blue sleep area. by analogy, the bed with blue sleep area has the 
same positional relation to an object of what category?
Answer:

\end{verbatim}

\section{Experiment on Low Resource Setting}
We observed that under max\_step of 30, our method outperforms on most questions types, but for the Long Rel type, our method is a little lower.
It is worth noting that our method significantly outperforms the naive ToT method when the maximum step size threshold is reduced to a smaller bound, say 10 or 20. This observation serves to demonstrate the efficacy of our approach in low resource scenarios.

\begin{table*}[!ht]
    \centering
    \setlength{\tabcolsep}{1.1mm}
    \begin{tabular}{clccccccccccccccc}
    \toprule
    & & \multicolumn{3}{c}{\bf ToT} & & \multicolumn{3}{c}{\bf ToT-OS} & & \multicolumn{3}{c}{\bf ToT-block}  \\
    \cmidrule{3-5} \cmidrule{7-9} \cmidrule{11-13}
    & {max step} & 10 & 20 & 30 & &  10 & 20 & 30 & & 10 & 20 & 30 \\
    \midrule
    \multirow{2}{*}{
        \rotatebox[origin=c]{0}{{\it Sequence}}
    } & Short Rel & 50.55 & 52.33  & 52.66 & & 56.33 & 58.00 & 58.33 & & 57.33 & 59.00 & 59.33 \\
      & Long Rel  &  19.00 & 46.33 & 56.00 & & 38.00 & 48.33 & 53.00 & & 31.00 & 49.66 & 55.00 \\
    \midrule
    \multirow{3}{*}{
        \rotatebox[origin=c]{0}{{\it Parallel}}
    } & Sum     & 93.66 & 98.33 & 98.66 & & 99.33 & 99.66 & 100.00 & & 99.66 & 100.00 & 100.00 \\
      & Compare & 57.00 & 59.33 & 59.66 & & 63.00 & 63.00 & 63.00 & & 60.66 & 61.00 & 61.33 \\
      & Logic   & 34.66 & 58.66 &  65.00 & & 57.66 & 63.33 & 69.00 & & 62.33 & 68.33 & 70.66 \\
    \midrule
    \multirow{3}{*}{
        \rotatebox[origin=c]{0}{{\it Backtrack}}
    } & Query Part & 77.00 & 77.00 & 77.00 & & 78.66 & 78.66 & 78.66 & & 78.33 & 78.33 & 78.33 \\
      & Exist      & 74.00 & 79.66 & 82.00 & & 82.00 & 88.66 & 90.33 & & 78.00 & 85.00 & 86.66 \\
      & Count & 76.66 & 83.33 & 85.66 & & 82.66 & 88.00 & 90.00 & & 85.00 & 90.33 & 92.00 \\
    \midrule
    \multirow{1}{*}{
    } 
    \makecell{
        {\it Multi} \\ {\it Backtrack}
    } & Analogy & 78.00 & 88.00 & 89.66 & & 89.66 & 92.00 & 93.00 & & 90.00 & 91.66 & 92.33 \\ 
         \bottomrule
    \end{tabular}
    \caption{The Accuracy under different max\_step setting.}
    \label{tab:alantion}
\end{table*}



\end{document}